\pdfoutput=1

\documentclass[11pt]{article}

\usepackage[preprint]{acl}

\usepackage{times}
\usepackage{soul}
\usepackage{latexsym}
\usepackage{booktabs}
\usepackage{cleveref}
\usepackage{xspace}
\usepackage{comment}
\usepackage[]{mdframed}
\usepackage{tcolorbox}
\usepackage{booktabs}

\newcommand{\bftab}{\fontseries{b}\selectfont}
\usepackage[colorinlistoftodos,prependcaption,textsize=tiny]{todonotes}
\usepackage[T1]{fontenc}

\usepackage[utf8]{inputenc}

\usepackage{microtype}

\usepackage{inconsolata}

\usepackage{graphicx}

\newcommand{\directid}{\textsc{details}\@\xspace} 
\newcommand{\family}{\textsc{family}\@\xspace}  
\newcommand{\body}{\textsc{appearance}\@\xspace}   
\newcommand{\details}{\textsc{circumstances}\@\xspace}   
\newcommand{\socio}{\textsc{sec}\@\xspace} 
\newcommand{\fclt}{\textsc{fclt\_personnel}\@\xspace}   
\newcommand{\reltime}{\textsc{time}\@\xspace} 
\newcommand{\lfstl}{\textsc{lfstl}\@\xspace}  
\newcommand{\other}{\textsc{other}\@\xspace}   

\title{Beyond De-Identification: A Structured Approach for Defining and Detecting Indirect Identifiers in Medical Texts}

\author{Ibrahim Baroud$^{1,2}$, Lisa Raithel$^{1,2,3}$, Sebastian Möller$^{1,2}$, Roland Roller$^{1,2}$  \\
  $^1$Quality \& Usability Lab, Technische Universität Berlin \\
  $^2$German Research Center for Artificial Intelligence (DFKI), Berlin \\
  $^3$BIFOLD – Berlin Institute for the Foundations of Learning and Data\\
  \texttt{\{ibrahim.baroud@tu-berlin.de\}}\\}

\begin{document}
\maketitle
\begin{abstract}
Sharing sensitive texts for scientific purposes requires appropriate techniques to protect the privacy of patients and healthcare personnel. Anonymizing textual data is particularly challenging due to the presence of diverse unstructured direct and indirect identifiers. To mitigate the risk of re-identification, this work introduces a schema of nine categories of indirect identifiers designed to account for different potential adversaries, including acquaintances, family members and medical staff. Using this schema, we annotate 100 MIMIC-III discharge summaries and propose baseline models for identifying indirect identifiers. We will release the annotation guidelines, annotation spans (6,199 annotations in total) and the corresponding MIMIC-III document IDs to support further research in this area. 
\end{abstract}

\section{Introduction}

Access to data remains a major bottleneck in developing machine learning models for healthcare. 
Since data contains sensitive details about individuals, it cannot be shared readily outside hospitals. 
Interactions with legal departments and data security can be cumbersome, and regulations are somewhat unclear, particularly where text is concerned. 
However, the concept of de-identification is well-defined: According to HIPAA,\footnote{The U.S. Health Insurance Portability and Accountability Act of 1996.} it requires the removal of a list of direct identifiers, known as protected health information (PHI)\footnote{\url{https://www.hhs.gov/hipaa/for-professionals/special-topics/de-identification/index.html}}, including, e.g., names and addresses.

Classical de-identification of text data has been explored for many years with various approaches \cite{sweeney1996replacing,gupta2004evaluation,he2015crfs,kocaman2023rwd143}
\begin{figure}[ht]
\small
    \centering
    \begin{tcolorbox}[colframe=green!25, colback=green!10]

[...] Patient is a \textcolor{red}{\textit{33-year-old}} male, admitted at \textcolor{red}{\textit{12:20}} after a \textcolor{red}{\textit{motor vehicle accident}}.

[...] He \textcolor{red}{\textit{works as a carpenter}} and \textcolor{red}{\textit{lives with his 28-year-old girlfriend in assisted living}}. No known \textcolor{red}{\textit{health insurance}}, and he is \textcolor{red}{\textit{currently on disability assistance}}. [...] He was noted to be \textcolor{red}{\textit{obese (BMI 32)}} with a \textcolor{red}{\textit{height of 178 cm}} and \textcolor{red}{\textit{weight of 110 kg}}. 

[...] He was evaluated by the \textcolor{red}{\textit{Emergency Department team}} and consulted with \textcolor{red}{\textit{Orthopedics}} for suspected fractures. [...] Patient reports \textcolor{red}{\textit{playing basketball once a week}} [...].
    \end{tcolorbox}
    \caption{A snippet of a fictitious discharge summary with annotations according to our IPI schema in red.}
    \label{fig:intro_ex}
\end{figure}
and state-of-the-art de-identification systems achieve an $F_1$-score $\geq 95\%$ on academic benchmarks \cite{kocaman2023rwd143, yogarajan2020review}.
However, additional manual effort is needed to remove remaining PHIs, and more importantly, unstructured text often contains \textbf{additional information beyond PHIs} that can reveal an individual's identity \cite{feder2020active}, making the manual inspection process even more complex.

The concept of anonymization goes further: It is defined as an irreversible procedure that is applied to the data such that no information can be linked to any specific individual anymore \cite{meystre2010automatic}.
While the terms de-identification and anonymization are often used interchangeably, they refer to distinct concepts \cite{chevrier2019use}. 
De-identification focuses solely on removing direct identifiers, whereas anonymization must also address indirect identifiers.
Indirect identifiers are pieces of information that are potentially publicly known about an individual but do not lead to reidentification when considered alone.
However, in combination with other background or external knowledge, they can be used to uniquely identify an individual \citep{coli_a_00458}. 
\Cref{fig:intro_ex} shows a synthetic discharge summary with highlighted information (beyond direct identifiers) that may help reveal the identity of a person.

Despite the importance of anonymization, relatively few studies have systematically addressed text anonymization beyond traditional PHI detection. \citet{gardner2008hide} developed a system for extracting and suppressing sensitive information other than PHIs, but limited to diagnoses. 
\citet{kolditz2019annotating} created a dataset with PHIs and added more categories, namely medical units, relatives, and typists.  
\citet{feder2020active} annotated a set of demographic traits in clinical notes and proposed a framework for detecting sentences that include such traits.
\citet{coli_a_00458} presented a benchmark comprising annotations of court cases and evaluation metrics to assess the performance of anonymization methods. 
The annotations cover categories such as names
and quantities, 
and annotators mark each of the entities as a direct or indirect identifier. 
Moreover, \citet{yang2024robust} proposed a framework for text anonymization 
based on large language models (LLMs).
This framework measures anonymization success by only checking whether an adversarial LLM can guess the name of the person to which the text belongs.

Building prior work, our study defines and identifies information beyond traditional personal health identifiers within a controlled framework.
We introduce
a schema of indirect personal identifiers (IPIs) optimized for a medical context and apply it to annotate relevant spans in discharge summaries from the Medical Information Mart for Intensive Care (MIMIC-III) dataset \citep{johnson2016mimic}. We define the problem of structurally identifying IPIs as a span classification problem rather than a sentence classification problem as done by \citet{feder2020active} to avoid removing whole sentences (which might include other medical related information) and reduce information loss during anonymization. 
Finally, we evaluate the performance of various models in detecting these annotated identifiers.

\section{Indirect Personal Identifiers (IPI)}

The type of information that may lead to re-identification in a given text is domain-dependent and requires unique analysis \citep{sweeney2000simple}.
In the following, we introduce a schema covering aspects of indirect personal identifiers (IPI) and apply it to annotate spans in discharge summaries from MIMIC-III. 
To construct our dataset, we randomly sampled 100 summaries with a length ranging from 500 to 2,500 words.\footnote{More details on the dataset in \Cref{app:annotations}.}

\subsection{IPI Schema} \label{categorization}

Our proposed schema builds on related work by \citet{kolditz2019annotating} and \citet{feder2020active}, as well as our manual analysis of discharge summaries. 
From prior work, we incorporate concepts like \textit{medical unit} \citep{kolditz2019annotating}, expanding it to include medical services, teams and medical personnel. 
We adapt \textit{family structure} from \citet{feder2020active}, broadening it to include family decisions and integrate \textit{living arrangements} into a new category, \directid, which covers indirect identifiers such as addresses (e.g., `lives in prison'), dates (`he turned 18 right before COVID started'), and references to other PHIs like licence numbers. 

Additionally, we adapt the category \textit{occupation} into \socio, which covers socio-economic and criminal history.
Our \lfstl category includes habits, sports, and diet, alongside the \textit{drug} category from \citet{feder2020active}. 
We redefine the category \textit{casually noticeable} in our category \body to specifically cover body piercings, tattoos and scars.
Based on our manual analysis, we introduce \reltime to capture time-related expressions such as timestamps for taking lab values, admission days, and time references around events such as surgeries.  
A brief overview of our final categories is provided below,\footnote{The following examples were created by the authors to avoid presenting data from MIMIC-III directly.} with further details available in \Cref{sec:details_categories}.

\begin{description}
  \item[\body] Descriptions of appearance, e.g. \textit{freshly healed scar behind right ear}, and mentions of weight, height, or body modifications.
  
  \item[\details] Any mention of an event (e.g. an accident) that caused an injury or happened in a medical facility.
  This category also includes specific statements or behavior, e.g. \textit{crashed his car into a dumpster} or \textit{refused medication because she does not believe in it}.
  
  \item[\socio] Mentions of information concerning \underline{s}ocio-\underline{e}conomic or \underline{c}riminal history, such as employment (e.g. \textit{is a retired police officer}), health insurance (e.g. \textit{has no health insurance}), or social/legal status (\textit{does not have valid papers}).
  
  \item[\family] Any mention of family-related information, such as being adopted, as well as the family's medical history or involvement (e.g. \textit{daughter serves as her health care proxy}). 
  
  \item[\fclt] Mentions of healthcare facilities (\textit{ICU}) or medical personnel (\textit{nursing team}).

  \item[\reltime] All mentions of age or time-related information (e.g., \textit{postoperative day number 5}).

  \item[\lfstl] Regular activities and habits, such as sports or diet (e.g. \textit{reports sticking to low-sodium diet}), but also tobacco, alcohol, or substance use. 

  \item[\directid] 
  All mentions of PHIs that were not detected, or a description of a PHI (e.g. \textit{lives in a halfway house}, which reveals information about the person's address).
  
  \item[\other] All other kinds of non-medical but infrequent information that might be sensitive
  , e.g. languages, ethnicity, or sexual orientation. 
  
\end{description}

\subsection{Data Annotation}
Two annotators independently labelled the same set of 100 de-identified discharge summaries using the nine categories described above.\footnote{Annotation guidelines will be published upon acceptance.}
The annotations were then consolidated,
and inter-annotator agreement (IAA) was calculated using the average pairwise relaxed $F_1$-score between the annotators' marked entities \citep{hripcsak2005agreement}.\footnote{Details about the annotators and IAA can be found in \Cref{app:annotations}.}
The overall agreement resulted in an $F_1$-score of $0.87$. 
\Cref{tab:iaa} in \Cref{app:annotations} lists the scores for each category. 
The annotators achieved the highest agreement in the categories \reltime ($F_1=0.89$), \lfstl ($F_1=0.88$) and \family ($F_1=0.87$), and the lowest on \directid ($F_1=0.41$).


The finalized dataset consists of 6,199 annotations, the majority of them belong to the categories \reltime (64.62\%) and \fclt (22.92\%). 
This is expected, as most discharge summaries contain detailed temporal descriptions, department consultations, and precise timestamps, such as when lab values were recorded. In contrast, information such as spoken languages or accident details appeared less frequently, as they were case-dependent and varied based on the typist's preference. 
\Cref{tab:annotations_stats} shows the number of annotations per category and their percentage in the overall annotations. 

\begin{table}[h]
\centering
\small
\begin{tabular}{lrr}
\toprule
Category       & \#Annotations & Proportion \\
\midrule
\family         & 273            & 4.4\%      \\
\body     & 132            & 2.13\%     \\
\details        & 99             & 1.6\%      \\
\socio          & 59             & 0.95\%     \\
\fclt   & 1421           & 22.92\%    \\
\reltime & 4006           & 64.62\%    \\
\lfstl          & 144            & 2.32\%     \\
\directid    & 32             & 0.52\%     \\
\other          & 33             & 0.53\%    \\
\bottomrule
\end{tabular}
\caption{Number of annotations per category in 100 discharge summaries from MIMIC-III.}
\label{tab:annotations_stats}
\end{table}

\subsection{Data Characteristics}

Overall, we focused on identifying indirect identifiers on span level, that may either be publicly known or describe a person’s status, behaviour, or appearance. 
Our final curated annotations reveal various such risks. 
For example, spans labeled as \details contain descriptive information about accidents that could facilitate re-identification by witnesses.
These details may enable an adversary to retrieve additional information about the patient, e.g. by searching online to find reports about the incident. 
Moreover, this category might encompass other sensitive descriptions, such as instances of patient aggression toward staff or refusal of medication. 

The 59 annotations from the \socio category reveal information about a person's criminal history, which is public information in the U.S. \cite{jacobs2012criminal}, and therefore easy to look up even for a layperson.
This category covers mentions of the patient being incarcerated, which may, in some cases, reveal the exact address.
Finally, the annotations include various information about patients' social status, such as being homeless or not having health insurance, or lifestyle, such as information about drinking, smoking, or sports. 
Although these mentions are relatively infrequent in the dataset, they may pose a high re-identification risk. Unique or rare characteristics -- especially those that distinguish an individual from the broader population -- can drastically narrow down the pool of potential matches, making re-identification more feasible.

\section{Experiments}

To provide a first baseline for the automatic detection of the proposed set of indirect identifiers in medical texts, we experimented with BERT  \citep{Devlin2019BERTPO} as well as open-source LLMs.
We split the data into training (60\%), development (15\%), and test (25\%) sets, using the dev set for hyperparameter optimization. 
For span classification, we fine-tuned a BERT model using the HuggingFace library \cite{wolf-etal-2020-transformers}. 
For the LLM experiments, we used Llama-3.1-8b-Instruct, Mistral-7B-Instruct-v0.3, and Qwen2.5-14B-Instruct in both zero-shot and three-shot settings leveraging Declarative Self-improving Python (DSPy) \cite{khattab2023dspycompilingdeclarativelanguage}, to automatically refine and optimize the prompt.
Model performance was assessed using relaxed precision, recall, and $F_1$-score.
Further details on data preprocessing, model fine-tuning, and evaluation can be found in \Cref{sec:train_details}.

\subsection{Results}

Detailed evaluation results for the BERT model can be found in \Cref{tab:bert_resuls}. 
Notably, recall is higher than precision in almost all cases.
Phrases containing socio-economic or criminal information (\socio), medical facilities and personnel (\fclt), and time expressions (\reltime) achieve higher scores than the other categories, i.e., less frequent categories tend to have a lower $F_1$-score, which was also true for the IAA scores.
The lightweight LLMs, which are explored for the first time for this specific task, performed poorly on the test set with $F_1$-score $\leq 51\%$ (micro) and recall $\leq 47\%$ (see also \Cref{tab:all_resuls} and paragraph \textbf{LLMs} in \Cref{sec:train_details} for more information and discussion about the LLM results).
Our evaluation showed that the LLMs sometimes failed to follow the pre-defined output format and preserve the originality of the spans in the original texts. Moreover, they frequently hallucinated and extracted irrelevant or non-existent information. 

\begin{table}[h]
\small
\centering
\begin{tabular}{lrrrr}
\toprule
\textbf{Category}        & \textbf{P} & \textbf{R} & $\mathbf{F_1}$ & \textbf{Support} \\ \midrule
\directid & 0.13                                   & 0.50                                & 0.21                               & 4                \\
\family   & 0.67                                   & 0.96                                & 0.79                               & 73               \\
\body     & 0.52                                   & 0.59                                & 0.55                               & 29               \\
\details  & 0.18                                   & 0.23                                & 0.20                                & 30               \\
\socio    & 0.59                                   & 0.71                                & 0.65                               & 14               \\
\fclt     & 0.80                                   & 0.92                                & 0.85                               & 362              \\
\reltime  & 0.84                                   & 0.97                                & 0.90                                & 1006             \\
\lfstl    & 0.57                                   & 0.86                                & 0.68                               & 35               \\
\other    & 0.20                                   & 0.14                                & 0.17                               & 7                \\ \midrule
micro average            & 0.78                                   & 0.93                                & 0.85                               & 1560             \\
macro average            & 0.50                                   & 0.65                                & 0.55                               & 1560  \\ \bottomrule          
\end{tabular}
\caption{Evaluation results on the test set for the \textbf{BERT-based} system in \textbf{P}recision, \textbf{R}ecall, and $\mathbf{F_1}$ score. Support shows the number of examples in the test set.}
\label{tab:bert_resuls}
\end{table}

\section{Discussion}

As expected, the BERT-based model clearly outperformed the lightweight LLMs in both zero-shot and 3-shot settings, corroborating the results of \citet{naguib2024few} about BERT superiority against LLMs for span classification. This suggests that LLMs may be more powerful to support in validating anonymization systems through inferring hidden information as proposed by \citet{staab24beyond} rather than being used for span classification.

The BERT model shows a satisfactory micro $F_1$-score, with its comparably high recall being particularly advantageous for anonymization, as missing sensitive information can have serious consequences. 
However, the low macro $F_1$-score combined with the strong imbalance of the annotated categories indicates that the model struggles to detect less frequent, yet more critical, categories.

One reason for this may be the limited amount of training data, hampering the model to learn robust representations for rare categories.
Additionally, the inherent linguistic complexity within categories further complicates the task.
In contrast to PHIs, i.e., names or addresses, which usually follow similar patterns across documents, IPIs exhibit greater lexical and semantic diversity. 
This does not only make them more challenging, but also emphasizes the urgency of accurately identifying them for effective anonymization.
Given that annotating additional documents is both time- and resource-intensive, especially when rare events must be captured in sufficient numbers, it may be more realistic to investigate methods that perform well in low-resource scenarios. 

\section{Conclusion}

In this work, we introduced a dataset along with an annotation schema designed to capture a wide range of indirect identifiers in medical texts. 
The schema is inspired by medical records, but is adaptable to other domains and text genres with minimal modifications.
We evaluated the performance of BERT and LLMs in detecting the proposed categories.
The overall performance of the models highlights the inherent difficulty of this task, particularly in identifying less frequent and diverse indirect identifiers.
However, our work provides a foundation for further exploration and adaptation, with the goal to improve privacy through structural information detection.
In future work, we aim to develop a framework that (k-)anonymizes the proposed indirect identifiers and study the utility of the anonymized texts on downstream tasks.

\clearpage
\section{Limitations}

Our list of categories is diverse; however, indirect identifiers should not be limited to it, and further studies should explore more potential risks in unstructured data that do not fall under these categories. 
We plan to test the scalability of our schema to other datasets, languages and domains (such as legal or financial), but accessing similar relevant data is very limited due to privacy concerns, especially in languages other than English.

The LLM experiments are intended to provide a different baseline approach rather than to compare performance with the BERT model, as such a comparison would be unfair in a zero- or few-shot setting. The LLM approach could be improved, for example, by using bigger models or performing an instruction tuning using the training set instead of evaluating the models in a zero- or few-shot setting. We plan to use LLMs to augment the training set with synthetically generated examples to solve the problem of low number of examples for certain categories, which were also not enough to train the BERT model. 

BERT-based models have been shown to work well in NER tasks; however, they cannot be fully relied on for finding all mentions of potentially sensitive information.
Instead, these models can be used as a complement to help humans speed up the process of enhancing privacy. 
As for LLMs, we would not trust them to produce complete and reliable results since our experiments showed unfaithful output in terms of format (which hinders a structured evaluation) and ``hallucinations.'' 

We did not experiment with a hybrid approach (e.g., combining regular expressions and the approaches described) to improve the detection of categories with formulaic patterns for which we expect a better performance using regular expression such \reltime.

\section{Ethical Considerations}
The data used in the above work is publicly available, de-identified data from the MIMIC-III database and therefore does not expose any patients or medical staff.
It is only available after registration and training. 
We state that we only annotated potential indirect identifiers and did not attempt to re-identify any patients. 
All examples in this paper were created by the authors. 
They resemble texts from MIMIC-III, but are not copied from real discharge summaries. 
We will release the annotations and document IDs from MIMIC-III upon request, but \textbf{not} the documents themselves. 

\bibliography{custom}

\begin{thebibliography}{24}
\providecommand{\natexlab}[1]{#1}

\bibitem[{Chevrier et~al.(2019)Chevrier, Foufi, Gaudet-Blavignac, Robert, and Lovis}]{chevrier2019use}
Rapha{\"e}l Chevrier, Vasiliki Foufi, Christophe Gaudet-Blavignac, Arnaud Robert, and Christian Lovis. 2019.
\newblock {Use and understanding of anonymization and de-identification in the biomedical literature: scoping review}.
\newblock \emph{Journal of medical Internet research}, 21(5):e13484.

\bibitem[{Devlin et~al.(2019)Devlin, Chang, Lee, and Toutanova}]{Devlin2019BERTPO}
Jacob Devlin, Ming-Wei Chang, Kenton Lee, and Kristina Toutanova. 2019.
\newblock \href {https://api.semanticscholar.org/CorpusID:52967399} {Bert: Pre-training of deep bidirectional transformers for language understanding}.
\newblock In \emph{North American Chapter of the Association for Computational Linguistics}.

\bibitem[{Feder et~al.(2020)Feder, Vainstein, Rosenfeld, Hartman, Hassidim, and Matias}]{feder2020active}
Amir Feder, Danny Vainstein, Roni Rosenfeld, Tzvika Hartman, Avinatan Hassidim, and Yossi Matias. 2020.
\newblock Active deep learning to detect demographic traits in free-form clinical notes.
\newblock \emph{Journal of Biomedical Informatics}, 107:103436.

\bibitem[{Gardner and Xiong(2008)}]{gardner2008hide}
James Gardner and Li~Xiong. 2008.
\newblock {HIDE: an integrated system for health information DE-identification}.
\newblock In \emph{2008 21st IEEE international symposium on computer-based medical systems}, pages 254--259. IEEE.

\bibitem[{Gupta et~al.(2004)Gupta, Saul, and Gilbertson}]{gupta2004evaluation}
Dilip Gupta, Melissa Saul, and John Gilbertson. 2004.
\newblock Evaluation of a deidentification (de-id) software engine to share pathology reports and clinical documents for research.
\newblock \emph{American journal of clinical pathology}, 121(2):176--186.

\bibitem[{He et~al.(2015)He, Guan, Cheng, Cen, and Hua}]{he2015crfs}
Bin He, Yi~Guan, Jianyi Cheng, Keting Cen, and Wenlan Hua. 2015.
\newblock Crfs based de-identification of medical records.
\newblock \emph{Journal of biomedical informatics}, 58:S39--S46.

\bibitem[{Hripcsak and Rothschild(2005)}]{hripcsak2005agreement}
George Hripcsak and Adam~S Rothschild. 2005.
\newblock Agreement, the f-measure, and reliability in information retrieval.
\newblock \emph{Journal of the American medical informatics association}, 12(3):296--298.

\bibitem[{Jacobs and Larrauri(2012)}]{jacobs2012criminal}
James~B Jacobs and Elena Larrauri. 2012.
\newblock Are criminal convictions a public matter? the usa and spain.
\newblock \emph{Punishment \& Society}, 14(1):3--28.

\bibitem[{Johnson et~al.(2016)Johnson, Pollard, Shen, Lehman, Feng, Ghassemi, Moody, Szolovits, Anthony~Celi, and Mark}]{johnson2016mimic}
Alistair~EW Johnson, Tom~J Pollard, Lu~Shen, Li-wei~H Lehman, Mengling Feng, Mohammad Ghassemi, Benjamin Moody, Peter Szolovits, Leo Anthony~Celi, and Roger~G Mark. 2016.
\newblock Mimic-iii, a freely accessible critical care database.
\newblock \emph{Scientific data}, 3(1):1--9.

\bibitem[{Khattab et~al.(2023)Khattab, Singhvi, Maheshwari, Zhang, Santhanam, Vardhamanan, Haq, Sharma, Joshi, Moazam, Miller, Zaharia, and Potts}]{khattab2023dspycompilingdeclarativelanguage}
Omar Khattab, Arnav Singhvi, Paridhi Maheshwari, Zhiyuan Zhang, Keshav Santhanam, Sri Vardhamanan, Saiful Haq, Ashutosh Sharma, Thomas~T. Joshi, Hanna Moazam, Heather Miller, Matei Zaharia, and Christopher Potts. 2023.
\newblock \href {https://arxiv.org/abs/2310.03714} {Dspy: Compiling declarative language model calls into self-improving pipelines}.
\newblock \emph{Preprint}, arXiv:2310.03714.

\bibitem[{Kocaman et~al.(2023)Kocaman, Talby, and Hak}]{kocaman2023rwd143}
Veysel Kocaman, D~Talby, and H~Ul Hak. 2023.
\newblock {RWD143 Beyond Accuracy: Automated De-Identification of Large Real-World Clinical Text Datasets}.
\newblock \emph{Value in Health}, 26(12):S532.

\bibitem[{Kolditz et~al.(2019)Kolditz, Lohr, Hellrich, Modersohn, Betz, Kiehntopf, and Hahn}]{kolditz2019annotating}
Tobias Kolditz, Christina Lohr, Johannes Hellrich, Luise Modersohn, Boris Betz, Michael Kiehntopf, and Udo Hahn. 2019.
\newblock Annotating german clinical documents for de-identification.
\newblock In \emph{MEDINFO 2019: Health and Wellbeing e-Networks for All}, pages 203--207. IOS Press.

\bibitem[{Kwon et~al.(2024)Kwon, Kim, Lee, Bae, Kyung, Cha, Pollard, Johnson, and Choi}]{Kwon2024EHRConDF}
Yeonsu Kwon, Jiho Kim, Gyubok Lee, Seongsu Bae, Daeun Kyung, Wonchul Cha, Tom Pollard, Alistair Johnson, and Edward Choi. 2024.
\newblock \href {https://api.semanticscholar.org/CorpusID:270702556} {Ehrcon: Dataset for checking consistency between unstructured notes and structured tables in electronic health records}.
\newblock \emph{ArXiv}, abs/2406.16341.

\bibitem[{Meystre et~al.(2010)Meystre, Friedlin, South, Shen, and Samore}]{meystre2010automatic}
Stephane~M Meystre, F~Jeffrey Friedlin, Brett~R South, Shuying Shen, and Matthew~H Samore. 2010.
\newblock Automatic de-identification of textual documents in the electronic health record: a review of recent research.
\newblock \emph{BMC medical research methodology}, 10:1--16.

\bibitem[{Montani and Honnibal()}]{prodigy_montani_honnibal}
Ines Montani and Matthew Honnibal.
\newblock \href {https://prodi.gy/} {Prodigy: A modern and scriptable annotation tool for creating training data for machine learning models}.

\bibitem[{Naguib et~al.(2024)Naguib, Tannier, and N{\'e}v{\'e}ol}]{naguib2024few}
Marco Naguib, Xavier Tannier, and Aur{\'e}lie N{\'e}v{\'e}ol. 2024.
\newblock Few shot clinical entity recognition in three languages: Masked language models outperform llm prompting.
\newblock \emph{arXiv preprint arXiv:2402.12801}.

\bibitem[{Pilán et~al.(2022)Pilán, Lison, Øvrelid, Papadopoulou, Sánchez, and Batet}]{coli_a_00458}
Ildikó Pilán, Pierre Lison, Lilja Øvrelid, Anthi Papadopoulou, David Sánchez, and Montserrat Batet. 2022.
\newblock \href {https://doi.org/10.1162/coli_a_00458} {{The Text Anonymization Benchmark (TAB): A Dedicated Corpus and Evaluation Framework for Text Anonymization}}.
\newblock \emph{Computational Linguistics}, 48(4):1053--1101.

\bibitem[{Segura-Bedmar et~al.(2013)Segura-Bedmar, Mart{\'\i}nez, and Herrero-Zazo}]{segura-bedmar-etal-2013-semeval}
Isabel Segura-Bedmar, Paloma Mart{\'\i}nez, and Mar{\'\i}a Herrero-Zazo. 2013.
\newblock \href {https://aclanthology.org/S13-2056} {{S}em{E}val-2013 task 9 : Extraction of drug-drug interactions from biomedical texts ({DDIE}xtraction 2013)}.
\newblock In \emph{Second Joint Conference on Lexical and Computational Semantics (*{SEM}), Volume 2: Proceedings of the Seventh International Workshop on Semantic Evaluation ({S}em{E}val 2013)}, pages 341--350, Atlanta, Georgia, USA. Association for Computational Linguistics.

\bibitem[{Staab et~al.(2024)Staab, Vero, Balunović, and Vechev}]{staab24beyond}
Robin Staab, Mark Vero, Mislav Balunović, and Martin Vechev. 2024.
\newblock Beyond memorization: Violating privacy via inference with large language models.
\newblock In \emph{The Twelfth International Conference on Learning Representations}.

\bibitem[{Sweeney(1996)}]{sweeney1996replacing}
Latanya Sweeney. 1996.
\newblock Replacing personally-identifying information in medical records, the scrub system.
\newblock In \emph{Proceedings of the AMIA annual fall symposium}, page 333. American Medical Informatics Association.

\bibitem[{Sweeney(2000)}]{sweeney2000simple}
Latanya Sweeney. 2000.
\newblock Simple demographics often identify people uniquely.

\bibitem[{Wolf et~al.(2020)Wolf, Debut, Sanh, Chaumond, Delangue, Moi, Cistac, Rault, Louf, Funtowicz, Davison, Shleifer, von Platen, Ma, Jernite, Plu, Xu, Le~Scao, Gugger, Drame, Lhoest, and Rush}]{wolf-etal-2020-transformers}
Thomas Wolf, Lysandre Debut, Victor Sanh, Julien Chaumond, Clement Delangue, Anthony Moi, Pierric Cistac, Tim Rault, Remi Louf, Morgan Funtowicz, Joe Davison, Sam Shleifer, Patrick von Platen, Clara Ma, Yacine Jernite, Julien Plu, Canwen Xu, Teven Le~Scao, Sylvain Gugger, Mariama Drame, Quentin Lhoest, and Alexander Rush. 2020.
\newblock \href {https://doi.org/10.18653/v1/2020.emnlp-demos.6} {Transformers: State-of-the-art natural language processing}.
\newblock In \emph{Proceedings of the 2020 Conference on Empirical Methods in Natural Language Processing: System Demonstrations}, pages 38--45, Online. Association for Computational Linguistics.

\bibitem[{Yang et~al.(2024)Yang, Zhu, and Gurevych}]{yang2024robust}
Tianyu Yang, Xiaodan Zhu, and Iryna Gurevych. 2024.
\newblock Robust utility-preserving text anonymization based on large language models.
\newblock \emph{arXiv preprint arXiv:2407.11770}.

\bibitem[{Yogarajan et~al.(2020)Yogarajan, Pfahringer, and Mayo}]{yogarajan2020review}
Vithya Yogarajan, Bernhard Pfahringer, and Michael Mayo. 2020.
\newblock {A review of automatic end-to-end de-identification: Is high accuracy the only metric?}
\newblock \emph{Applied Artificial Intelligence}, 34(3):251--269.

\end{thebibliography}
\appendix

\section{Detailed Descriptions of the IPI Categories}
\label{sec:details_categories}

\begin{description}
  \item[\body] Mention of a person's (also infant's) weight, height, or a description of a person's body or body modifications, e.g., a scar under the eye, very tall, very short, gained/lost weight over a specific period of time, tattoos, piercings, etc.
  
  \item[\details] Any mention or description of an event (accident, storm, wildfire, etc.) that caused, e.g., a person's injury or happened in the clinical center such as patient being aggressive, rejecting help or medicine, leaving AMA (also discussions in the regard with persons outside the family) or injuring hospital staff. 
  Additionally, details about how the person was brought into the hospital or mentions of statements, requests or complaints expressed by the person.
  
  \item[\socio] Any mention of specific information about the person's employment (e.g., \textit{is a retired police officer})  or criminal history, health insurance (e.g., \textit{has no health insurance} or \textit{has a legal guard}) or social status such as homelessness or living in subsidized housing.
  
  \item[\family] All mentions of detailed family-related information about the person such as being adopted, having a twin sibling or having had a vitro fertilization pregnancy. 
  Furthermore, specific descriptions of the family's medical history (e.g., \textit{parent died at age 40}) or involvement (e.g., \textit{patient's daughter serves as her health care proxy}). 
  
  \item[\fclt] All mentions of hospital names, hospital units, labs, departments, facilities, consulting services/teams, floor and rooms, medical branches, outside doctors.
  
  \item[\reltime] Mentions of age or time-related information, e.g., \textit{postoperative day number 2}, \textit{day of delivery number 13}, \textit{day of life 6}, or exact mentions of times when lab values were taken, exact times about when medications should be taken. Do not consider times related to the disease itself, e.g., \textit{stopped breathing for 30 secs}. 
  
  \item[\lfstl] Hobbies and Lifestyle: such as sports or playing an instrument. Lifestyle: e.g., information about the patient's diet or private lifestyle.
  
  \item[\directid] All mentions of PHIs that were not detected and de-identified automatically or an abstract/indirect description of a PHI, for instance regarding address (e.g., \textit{lives in a halfway house} or \textit{lives in prison}). 
  Any information not related to PHIs such as weight or medical units are not part of this category and should be annotated as described in the other categories below. 
  For consistency, the following are the PHIs to consider for this category: Name, email addresses, geographic details, dates directly related to the individual, telephone, fax numbers, social security numbers, medical record numbers, health plan beneficiary numbers, account numbers, certificate and license numbers, vehicle and device identifiers, biometric identifiers and facial photograph, URL, IP addresses.

  \item[\other] Other kinds of non-medical information that might be sensitive to keep in the data e.g. languages, ethnicity (e.g., \textit{Caucasian}, \textit{AAF} etc.), and sexual orientation.
  
\end{description}

\section{Data and Annotation Details}\label{app:annotations}

\paragraph{Data}
The discharge summaries we use for demonstrating our schema are randomly sample from the Medical Information Mart for Intensive Care (MIMIC-III) dataset \citep{johnson2016mimic}.
I comprises health-related data from over 40,000 patients who stayed in critical care units of the Beth Israel Deaconess Medical Center between 2001 and 2012. 
Among other types of data, such as patient demographics, the database also includes various types of textual data, such as diagnostic reports and discharge summaries.  
We chose discharge summaries for our study, since these are richer in information than other notes in MIMIC-III.

\paragraph{Annotation Tool} For annotation, we used Prodigy \citep{prodigy_montani_honnibal}, version 1.11.11. 
It was run on a secure, lab-internal server; access was only permitted to the authors.

\paragraph{Annotators} 
The annotation team included one female and one male researcher, each with a different cultural background. 
Both annotators are fluent in English, though it is not their native language. 
One has expertise in computer science and data anonymization, and the other has experience in biomedical natural language processing. 
Neither has formal medical training, but both have experience in computational research and have contributed to various annotation projects in a research setting.
Both annotators were compensated as part of their regular researcher roles.

\paragraph{Inter-Annotator Agreement} The reported pairwise $F_1$-score is based on partial matches: A true positive exists when the compared spans overlap with at least one token and have the same label. 
We focus on partial matches because the exact span is not as important as in other entity recognition tasks, the main difficulty lies in finding the relevant information and removing it—anonymizing a longer span does not hurt the patient.

\begin{table}[h]
\centering
\begin{tabular}{lr}
\toprule
\textbf{Category}       & $\mathbf{F_1}$\textbf{-Score} \\
\midrule
\directid     & 0.41     \\
\family         & 0.87     \\
\body     & 0.62     \\
\details        & 0.59     \\
\socio          & 0.78     \\
\fclt   & 0.85     \\
\reltime & 0.89     \\
\lfstl          & 0.88     \\
\other          & 0.52     \\
\midrule
micro average  & 0.87     \\
macro average  & 0.71    
\end{tabular}
\caption{Inter-annotator agreement overall and per category using partial match pairwise $F_1$-scores \cite{hripcsak2005agreement}.}
\label{tab:iaa}
\end{table}

\section{Model Training and Evaluation Details}\label{sec:train_details}

\paragraph{Data Preprocessing}
In order to train an NER model, we converted the Prodigy annotations (each represented with a span start and end) to word-level annotations. 
Words annotated as part of a category received label prefixes \texttt{B} when they are at the beginning of a category, \texttt{I} when they lie within the category, and finally, words that were not part of any category received the label \texttt{O} (out). 
Since BERT cannot handle sequences longer than 512 sub-tokens, we split the discharge summaries into sections to avoid truncation and information loss. 
Prodigy's annotation output is already pre-tokenized and we used the pre-trained BERT-base-cased tokenizer for subword tokenization. 

We split the data into training (60\%), development (15\%), and test (25\%) sets and used the development set for hyperparameter optimization. 
\Cref{tab:split_stats} shows the statistics of the final data split. 

\begin{table}[h]
\centering
\begin{tabular}{lrrrr}
\toprule
      & \textbf{train} & \textbf{dev} & \textbf{test} & \textbf{total} \\ \midrule
\#documents   & 60                                  & 15                                & 25                                 & 100                                 \\ 
\#sections    & 592                                 & 162                               & 253                                & 1007                                \\ 
\#annotations & 3712                                & 927                               & 1560                               & 6199                                \\ \bottomrule
\end{tabular}
\caption{Statistics for the train, development, and test sets. `\#sections' represent the number of sections the documents were split into for each set.}
\label{tab:split_stats}
\end{table}

\paragraph{BERT Fine-Tuning}

For choosing the hyperparameters, a bert-base-cased model\footnote{\url{https://huggingface.co/google-bert/bert-base-cased}} was fine-tuned for maximally 15 epochs (early stopping after two epochs' patience) on the training set and evaluated on the development set using a grid search over learning rate values (1e-5, 2e-5, 3e-5, 4e-5, 5e-5) and batch size values (4, 8, 16).
After selecting the hyperparameters, we trained a BERT model on 75\% of the data (training and development combined) using the best-performing hyperparameters: 8 epochs, 3e-5 as the learning rate, and 8 as the batch size.

\paragraph{LLMs}
\label{Discussion_LLMs}
For the experiments using LLMs, we utilized the DSPy framework to extract the defined categories using different open-source LLMs in a zero-shot and few-shot settings. We implemented an LLM agent for each category and provided DSPy the desciption of each category as defined in the annotation guidelines. 
Additionally, we utilized the Pydantic \footnote{\url{https://pypi.org/project/pydantic/}} python library to obtain structured and type-validated output from the LLMs. 

As \Cref{tab:all_resuls} in \Cref{sec:train_details} shows, Qwen2.5-14B achieved the highest $F_1$-score between the evaluated LLMs. Mistral-7B-v0.3 and the 8-bit quantized version of Qwen2.5-72B achieved the highest recall. Whereas, the highest precision was achieved by Qwen2.5-14B, which was equal in both zero and 3-shot settings. The 3-shot setting did not always improve the performance. Interestingly, the performance dropped in some cases when providing the models with examples. A phenomenon that was also observed in \citet{Kwon2024EHRConDF} when using Llama3 for information extraction, where the model achieved in some cases better results in the zero-shot setting in comparison to few-shot. This and the overall low performance of the LLMs in comparison to BERT highlights the question about the suitability and effectiveness of using LLMs for extracting our proposed categories of indirect identifiers.

\paragraph{Evaluation Details}

We evaluated on the held-out test set using the \texttt{nervaluate} package,\footnote{\url{https://github.com/MantisAI/nervaluate}} which is a Python implementation for evaluating NER models as defined in the SemEval 2013 - 9.1 task \cite{segura-bedmar-etal-2013-semeval}. 
We report the results following the type evaluation schema, which requires some overlap between the system-tagged entity and the gold-standard annotation.

\begin{table}[h]
\small
\centering
\begin{tabular}{lrrrr}
\toprule
\textbf{Model}              & \textbf{P} & \textbf{R} & $\mathbf{F_1}$ & \textbf{Support} \\ \midrule

Llama-3.1-8B      & 0.08               & 0.40            & 0.13        & 1560             \\
Llama-3.1-8B 3-shot     & 0.18               & 0.35            & 0.24        & 1560             \\
Mistral-7B-v0.3 & 0.17            & \bftab 0.47         & 0.25     & 1560   \\
Mistral-7B-v0.3 3-shot & 0.05            & 0.30         & 0.09     & 1560   \\
Qwen2.5-14B & \bftab 0.64            & 0.42         & \bftab 0.51     & 1560   \\
Qwen2.5-14B 3-shot & \bftab 0.64            & 0.28         & 0.39     & 1560   \\
Qwen2.5-72B$^*$ & 0.48            & \bftab 0.47         & 0.48     & 1560   \\

\bottomrule
\end{tabular}
\caption{Micro-averaged test results for each LLM showing precision (\textbf{P}), recall (\textbf{R}), and $F_1$-score ($\mathbf{F_1}$). $^*$This is the 8-bit quantized version of this model. Values in \textbf{Bold} represent the highest performance for each metric among all tested LLMs}
\label{tab:all_resuls}
\end{table}

\paragraph{Use of AI Assistants} ChatGPT was partially used as an AI assistant for coding support.

\paragraph{Computing Environment}
The following packages were used for conducting the experiments:

\begin{itemize}
    \item Transformers version 4.44.2\footnote{\url{https://huggingface.co/}}
    \item spacy version 3.7.5\footnote{\url{https://spacy.io/}}
    \item Prodigy version 1.11.11\footnote{\url{https://prodi.gy/}}
\end{itemize}

The BERT experiments were run on a T4 GPU with 16GB. The LLMs were run on 2x NVIDIA RTX A6000 with 48GB each.
\clearpage
\onecolumn

\section{Example Annotation}\label{sec:ex_anno}

\begin{figure}[h]
    \centering
    \includegraphics[width=0.82\linewidth]{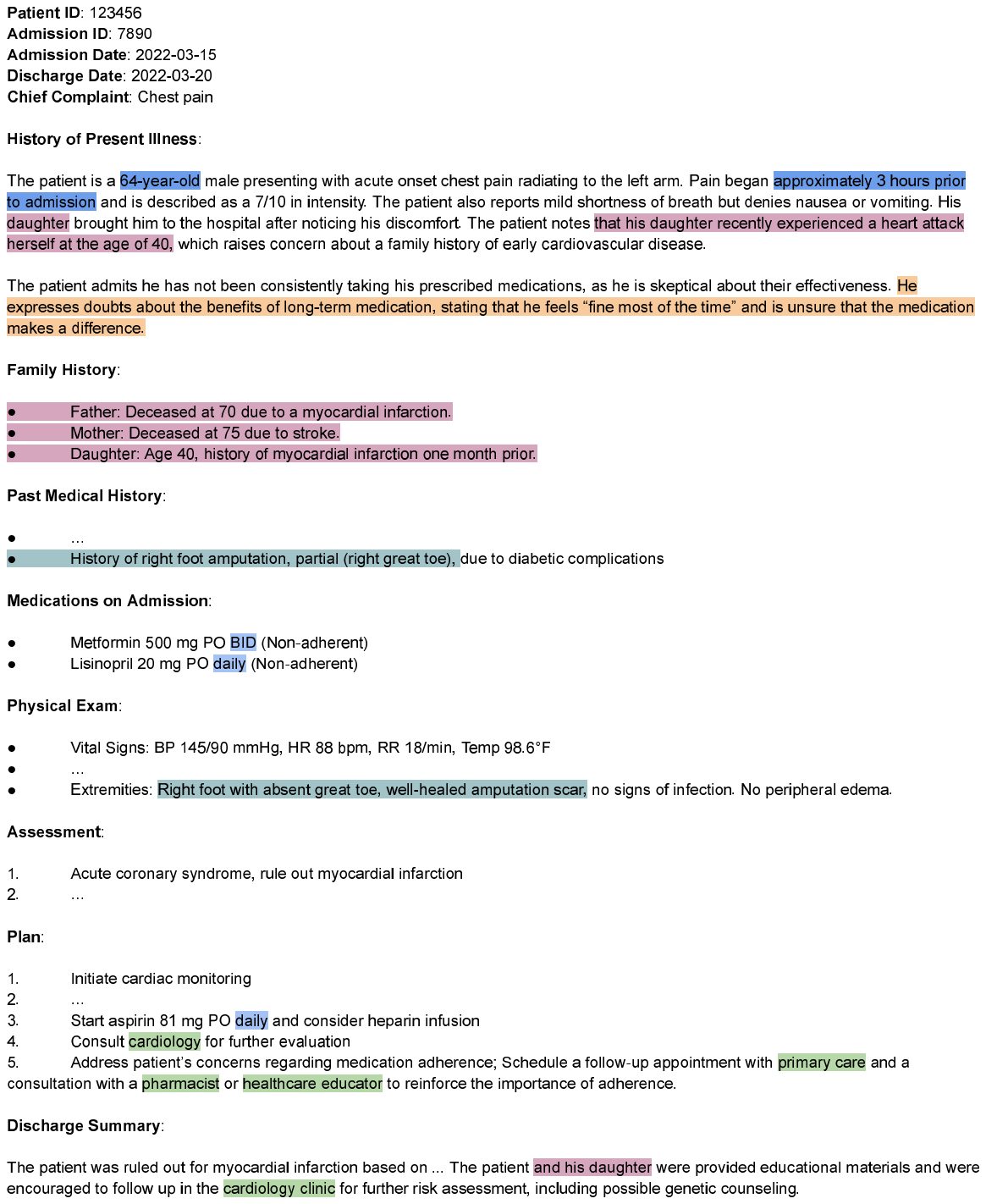}

    \caption{A (generated) discharge summary with annotations based on the
 proposed schema.}
    \label{fig:fake_ds}
\end{figure}

 \Cref{fig:fake_ds} shows an example of how the discharge summaries were annotated.

\clearpage
\onecolumn

\end{document}